\documentclass[lettersize,journal]{IEEEtran}
\usepackage{amsmath,amsfonts}
\usepackage{algorithmic}
\usepackage{algorithm}
\usepackage{array}
\usepackage[caption=false,font=normalsize,labelfont=sf,textfont=sf]{subfig}
\usepackage{textcomp}
\usepackage{stfloats}
\usepackage{url}
\usepackage{verbatim}
\usepackage{graphicx}
\usepackage{cite}
\usepackage{multirow}
\usepackage{booktabs}
\usepackage{xcolor}

\usepackage{colortbl} %
\usepackage{makecell}
\usepackage{amssymb}

\begin{document}

\title{SemGuard: Semantics-Decoupled Watermarking for Gaussian Splatting}

\author{Yingying Shi, Zhong Zhou, Bin Zhou
\thanks{Yingying Shi and Zhong Zhou are with the State Key Laboratory of Virtual Reality Technology and Systems, School of Computer Science and Engineering, Beihang University, Beijing, China, and also with Zhongguancun Laboratory, Beijing, China (e-mail: yyshi@buaa.edu.cn; zz@buaa.edu.cn).}%
\thanks{Bin Zhou is with the State Key Laboratory of Virtual Reality Technology and Systems, School of Computer Science and Engineering, Beihang University, Beijing, China, and also with the Peace-Bay Academy of Virtual Reality, Shenyang, China (e-mail: zhoubin@buaa.edu.cn).}%
\thanks{Corresponding author: Bin Zhou (e-mail: zhoubin@buaa.edu.cn).}%
}

\maketitle

\begin{figure*}[t!]
	\centering
 \includegraphics[width=7in]{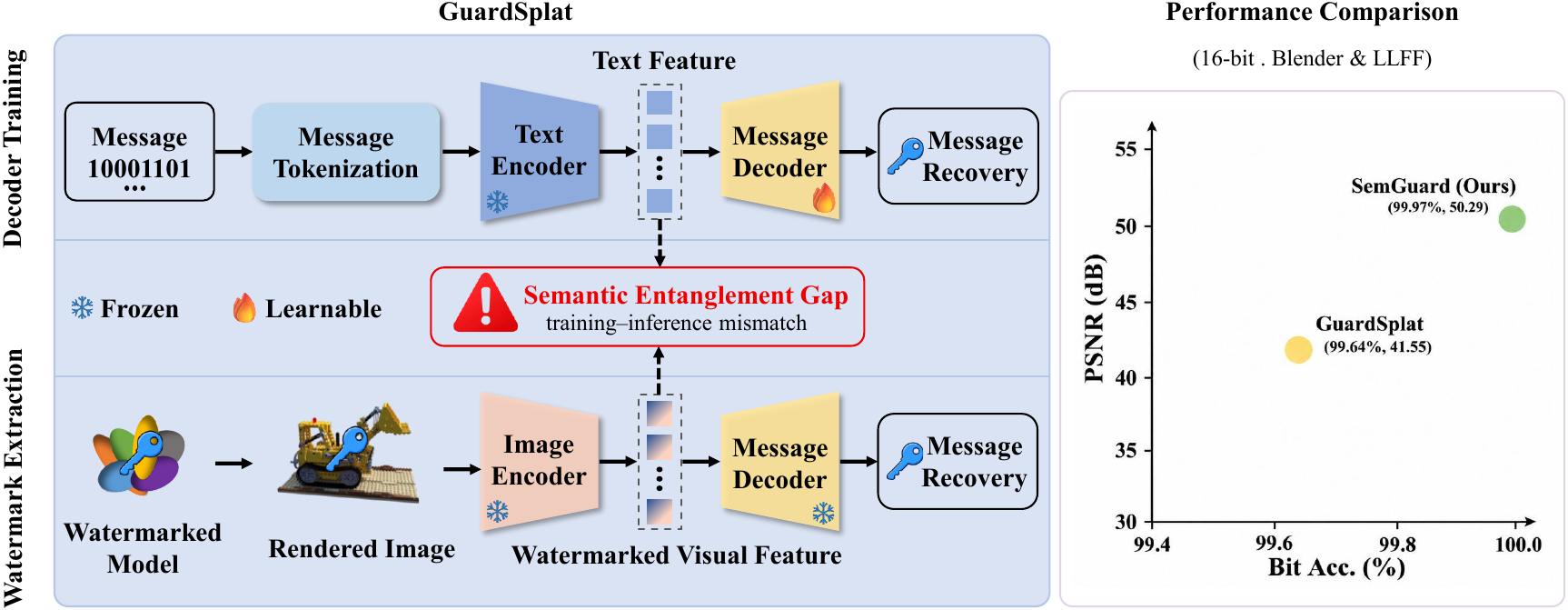}
\caption{
Motivation and performance comparison of SemGuard. GuardSplat pretrains a universal message decoder using CLIP text features but applies it to visual features containing both watermark information and scene-dependent semantics during extraction, resulting in a semantic discrepancy between decoder pretraining and watermark extraction. 
The right panel compares 16-bit watermark extraction accuracy and PSNR on Blender and LLFF, demonstrating that SemGuard substantially improves rendering fidelity while maintaining near-perfect watermark extraction accuracy.}
\label{fig:teaser}
\end{figure*}

\begin{abstract}
Gaussian Splatting provides an efficient representation for high-fidelity rendering of static and dynamic 3D scenes, motivating the copyright protection of Gaussian assets.  
In CLIP-guided Gaussian watermarking, a universal message decoder pretrained on textual watermark representations is subsequently applied to visual features containing both watermark information and scene-dependent semantics. 
However, text-only pretraining does not explicitly account for the semantic interference encountered during watermark extraction.
To address this challenge, we propose SemGuard, a simple yet effective semantics-decoupled watermarking framework that learns to suppress scene-dependent semantic interference during universal message decoder pretraining. 
Specifically, SemGuard combines message-related CLIP text features with message-independent visual features extracted from unwatermarked scene images, constructing training representations with controllable semantic interference relevant to the target Gaussian assets.
A progressive semantic interference curriculum gradually increases the contribution of visual features, allowing the decoder to first learn reliable message representations and then adapt to increasingly strong scene-dependent semantic interference.
After pretraining for a given message length and semantic image pool, the decoder is frozen and reused across multiple Gaussian assets whose visual semantics are covered by the pretraining data.
Experiments on static and dynamic Gaussian Splatting assets demonstrate improved rendering fidelity alongside reliable watermark extraction, while evaluations on static assets further show enhanced robustness against image-space distortions. 
Code is available at \url{https://github.com/shisyy/SemGuard}.

\end{abstract}



\section{Introduction}
\label{sec:introduction}

3D Gaussian Splatting (3DGS) has emerged as an efficient representation for high-fidelity novel-view synthesis and real-time rendering~\cite{kerbl20233d}. 
Extensions of Gaussian Splatting to dynamic scenes, including 4D Gaussian Splatting (4DGS), further enable the modeling and rendering of time-varying content~\cite{luiten2024dynamic,wu20244d,yang2024deformable}.
As Gaussian assets become increasingly accessible and widely shared, protecting their intellectual property and verifying ownership have become important concerns. 
Digital watermarking provides a practical solution by embedding imperceptible binary messages into digital content~\cite{cox1997secure,zhu2018hidden,tancik2020stegastamp}. Unlike watermarking individual images, protecting a Gaussian asset requires the embedded message to remain recoverable from different rendered observations of the same underlying representation. 
An effective Gaussian watermarking framework should therefore maintain reliable watermark extraction under varying viewpoints and temporal conditions while minimizing watermark-induced rendering distortions.

Recent studies have explored watermarking techniques for Gaussian Splatting assets~\cite{huang2024gaussianmarker,jang20253d,chen2025guardsplat}. 
Among them, GuardSplat~\cite{chen2025guardsplat} introduces an efficient CLIP-guided watermarking framework by pretraining a universal message decoder in the shared vision-language embedding space of CLIP~\cite{radford2021learning}.
Specifically, binary messages are converted into CLIP textual representations, from which a lightweight decoder learns to recover the corresponding message bits. After pretraining, the decoder is frozen and reused to extract watermark messages from CLIP visual features of rendered Gaussian views. 
However, a discrepancy remains between decoder pretraining and actual watermark extraction. During pretraining, the decoder learns message recovery solely from CLIP text features, whereas during extraction, it receives CLIP visual features containing both watermark-related information and scene-dependent semantics. Although both stages operate in CLIP's shared vision-language embedding space, text-only pretraining does not explicitly expose the decoder to the visual semantic interference encountered during watermark extraction.
We refer to this discrepancy between watermark-centric textual training features and semantics-entangled visual extraction features as the \emph{semantic entanglement gap}, which can hinder reliable watermark recovery when appearance modifications are constrained to preserve rendering fidelity, as illustrated in Fig.~\ref{fig:teaser}.

Addressing semantic interference while retaining decoder reusability remains an important challenge. Guard4D~\cite{shiguard4d} improves watermark extraction from dynamic Gaussian assets by introducing a temporal modeling and semantic decoupling (TMSD) module, which aggregates temporal information and suppresses scene-dependent semantic interference using control clips. Although its pretrained message decoder remains frozen, the TMSD module requires asset-specific optimization, introducing additional overhead into the complete watermark extraction pipeline.
This motivates a complementary approach to handling semantic interference: instead of learning an additional semantic correction module during asset-specific watermark optimization, can a message decoder learn to suppress the influence of scene-relevant visual semantics during pretraining and subsequently be reused across multiple Gaussian assets? Such a design shifts semantic-interference handling toward a shared pretraining stage, allowing the same frozen decoder to guide watermark embedding in different assets whose visual semantics are represented by the training image pool.

To this end, we propose SemGuard, a simple yet effective semantics-decoupled watermarking framework that incorporates visual semantic interference directly into universal decoder pretraining. 
Instead of learning message recovery solely from textual watermark representations, SemGuard constructs semantics-entangled training features by combining message-related CLIP text features with message-independent visual features extracted from unwatermarked images. 
The auxiliary visual features provide visual semantics relevant to the target Gaussian assets, while the binary message remains the supervision target. 
By supervising message recovery from these mixed representations, the decoder is encouraged to rely on watermark-related information while reducing its sensitivity to scene-dependent semantic interference, thereby learning to recover watermark messages in the presence of visual semantics represented during pretraining. 
Furthermore, we introduce a progressive semantic interference curriculum that gradually increases the contribution of visual features during pretraining, allowing the decoder to first establish a reliable message representation and then adapt to increasingly challenging semantic interference. 
For each message length and semantic image pool, SemGuard trains a single reusable message decoder, which is then frozen and shared by multiple Gaussian assets without further decoder adaptation. Each target asset is watermarked independently by optimizing only its spherical-harmonic appearance offsets, while the pretrained CLIP encoders and message decoder remain fixed. In this way, semantic-interference learning is performed jointly across a collection of scene semantics during decoder pretraining, whereas watermark embedding remains asset-specific and explicitly constrained to preserve the original rendered appearance.

Extensive experiments demonstrate substantial improvements in rendering fidelity while maintaining reliable watermark extraction, together with enhanced robustness against image-space distortions on static Gaussian assets. In particular, for 16-bit watermarking of static Gaussian assets, SemGuard improves PSNR from 41.55 dB to 50.29 dB and reduces LPIPS from 0.0017 to 0.0004 compared with GuardSplat, while maintaining a near-perfect Bit Acc. of 99.97\%.
Our main contributions are summarized as follows:

\begin{itemize}
\item We identify the lack of scene-dependent visual semantics during text-only decoder pretraining as a key challenge in CLIP-guided Gaussian watermarking, particularly when reliable watermark recovery must be achieved under rendering fidelity constraints.

\item We propose SemGuard, a simple yet effective semantics-decoupled watermarking framework that combines message-related CLIP text features with scene-relevant visual features under a progressive mixing curriculum. By learning message recovery in the presence of visual semantics covered by the pretraining pool, a single frozen decoder can be reused across multiple Gaussian assets without asset-specific decoder adaptation.

\item We evaluate SemGuard on static and dynamic Gaussian Splatting assets, demonstrating substantial improvements in rendering fidelity while maintaining reliable watermark recovery, 
together with enhanced robustness against image-space distortions on static assets.
\end{itemize}

\section{Related Work}
\label{sec:related_work}

\subsection{Watermarking across Visual and Neural Representations}
Digital watermarking has been extensively studied across different visual representations to protect content ownership while balancing imperceptibility and robustness.
Classical image watermarking methods typically embed information in spatial or transform domains~\cite{cox1997secure,navas2008dwt}, while learning-based approaches employ trainable embedding and extraction networks to improve robustness against image distortions and post-processing operations~\cite{fang2022end,wu2025robust,tancik2020stegastamp,fernandez2022watermarking,ma2022towards}.
Watermarking and information-hiding techniques have also been developed for generative visual content~\cite{fernandez2023stable,wen2023tree} and geometric representations, including meshes and point clouds~\cite{praun1999robust,ohbuchi2002frequency,zhu2021gaussian,zhang2023adaptive,zhu2024rethinking,yoo2022deep,zaman2025deep}. 
With the development of neural rendering~\cite{barron2021mip,barron2022mip,muller2022instant}, watermarking has further expanded to implicit scene representations, where messages are embedded into neural radiance fields and recovered from synthesized views~\cite{luo2023copyrnerf,jang2024waterf}. ZeroW-NeRF~\cite{zhu2026zerow} further explores neural scene copyright protection through an uncertainty-driven zero-watermarking strategy without modifying the original model parameters.
These studies demonstrate the applicability of watermarking and information hiding across visual, geometric, and neural scene representations, motivating further research on protecting explicit Gaussian assets. 

\begin{figure*}[t!]
\centering
\includegraphics[width=5.8in]{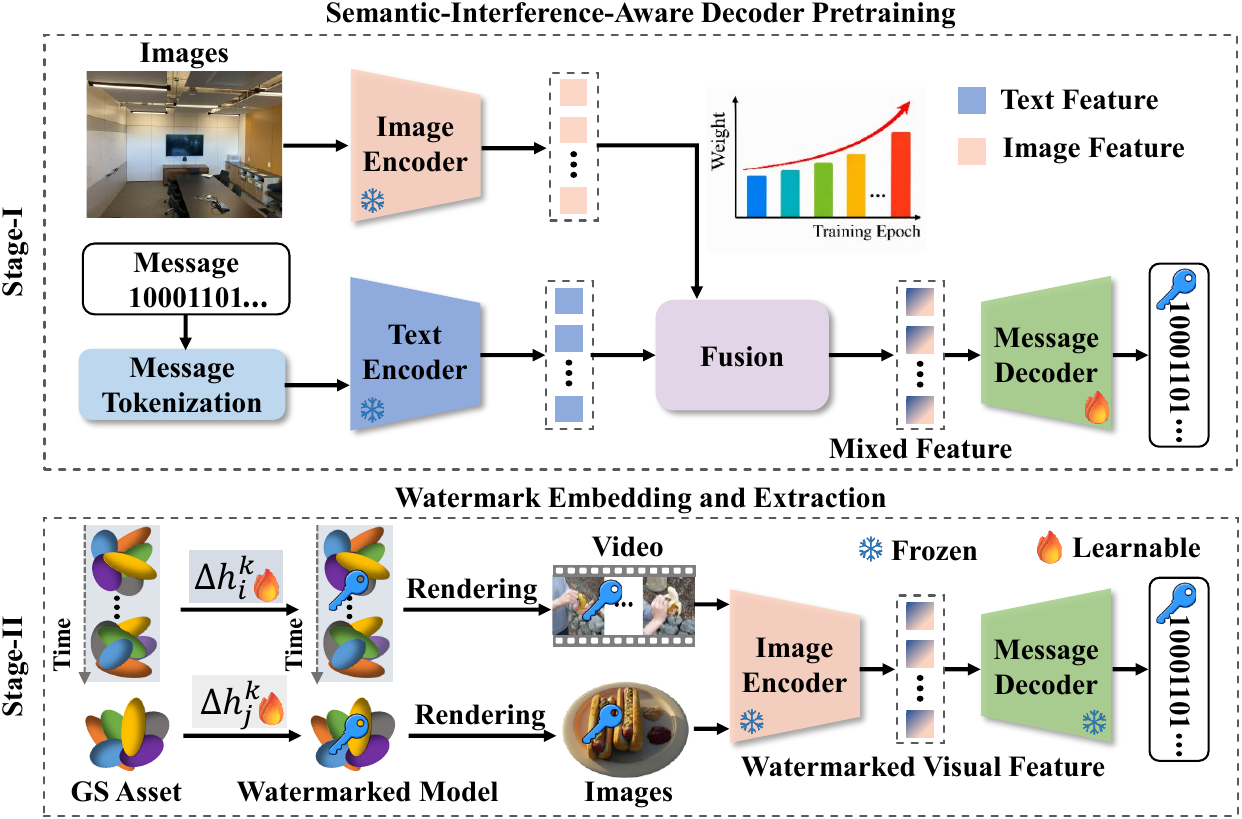}
\caption{Overview of SemGuard. 
In Stage I, message-related CLIP text features are mixed with visual features from unwatermarked scene images that cover the target Gaussian assets, and a progressive mixing curriculum is used to train a shared semantic-interference-aware message decoder. 
In Stage II, the pretrained decoder and CLIP encoders are frozen, while asset-specific spherical-harmonic offsets are optimized to embed the target message with minimal rendering distortion in static and dynamic Gaussian assets.}
\label{fig:method}
\end{figure*}

\subsection{Watermarking Static Gaussian Assets}
Recent advances in 3D Gaussian Splatting have improved rendering fidelity through adaptive spherical-harmonic modeling and multi-scale rendering strategies~\cite{huang2025structgs,zhao2026msa}.
The explicit and differentiable representation of 3D Gaussian Splatting~\cite{kerbl20233d,yu2024mip} enables watermark information to be embedded directly into Gaussian attributes. However, modifying these attributes may introduce visible rendering distortions, requiring a careful balance between watermark recoverability and visual fidelity. Existing methods address this challenge through different representation-aware embedding strategies. GaussianMarker~\cite{huang2024gaussianmarker} employs uncertainty-guided Gaussian densification to embed watermark information while limiting rendering distortions. 
GS-Hider~\cite{zhang2024gs} replaces the original SH coefficients with coupled secured features and employs separate scene and message decoders to recover rendered content and hidden information.
Meanwhile, 3D-GSW~\cite{jang20253d} combines frequency-guided densification, gradient masking, and wavelet-domain supervision to enhance rendering quality and robustness against image- and model-level distortions. 
Complementing these embedding-oriented approaches, GuardSplat~\cite{chen2025guardsplat} leverages the shared vision-language embedding space of CLIP~\cite{radford2021learning} to pretrain a universal message decoder using textual watermark representations, which is subsequently frozen for message extraction from Gaussian renderings.
Despite its reusable decoder design, GuardSplat relies on text-only pretraining without explicitly accounting for scene-dependent visual semantics encountered during watermark extraction, leaving the influence of semantic interference on the fidelity--recoverability trade-off insufficiently addressed.

\subsection{Watermarking Dynamic Gaussian Assets}
Compared with static Gaussian assets, dynamic Gaussian representations introduce additional challenges for watermark extraction due to temporal deformation and appearance variations~\cite{luiten2024dynamic,yang2024deformable,wu20244d,li2024spacetime}. These temporal changes further complicate reliable watermark recovery while preserving rendering fidelity.
Guard4D~\cite{shiguard4d} introduces a temporal modeling and semantic decoupling (TMSD) module that aggregates temporal information and mitigates semantic interference using clean and noise-perturbed control clips rendered from the unwatermarked asset. 
While its pretrained message decoder remains frozen, the TMSD module requires additional asset-specific optimization. 
Mark4D~\cite{lee2026mark4d} focuses on temporal consistency through watermark decoding in a latent video--text space, trajectory-aligned offsets, and motion-adaptive loss weighting.
Nevertheless, achieving reliable watermark recovery under temporal variations while preserving rendering fidelity and avoiding additional asset-specific semantic processing remains an important challenge.


\section{Method}
\label{sec:method}
As illustrated in Fig.~\ref{fig:method}, SemGuard consists of two stages: semantic-interference-aware decoder pretraining and SH-offset Gaussian watermark embedding. In Stage I, we combine message-related CLIP text features with message-independent visual features extracted from unwatermarked scene images, enabling the decoder to learn watermark recovery under scene-dependent semantic interference relevant to the target Gaussian assets. After pretraining for a given message length and semantic image pool, the decoder is frozen and reused across multiple Gaussian assets covered by the pretraining data. In Stage II, we optimize asset-specific SH offsets to embed the target watermark while keeping the Gaussian geometry, deformation parameters, CLIP encoders, and message decoder fixed. This design aims to improve rendering fidelity while maintaining reliable watermark extraction.

\subsection{Gaussian Splatting Preliminaries}
\label{sec:prelim}

A Gaussian Splatting asset is represented by a set of anisotropic Gaussian primitives~\cite{kerbl20233d}:
\begin{equation}
\mathcal{G}=\left\{(\boldsymbol{\mu}_i,\boldsymbol{\Sigma}_i,\alpha_i,\mathbf{h}_i)\right\}_{i=1}^{N},
\end{equation}
where $N$ denotes the number of Gaussians, and $\boldsymbol{\mu}_i$, $\boldsymbol{\Sigma}_i$, $\alpha_i$, and $\mathbf{h}_i$ represent the position, covariance, opacity, and spherical-harmonic (SH) appearance coefficients of the $i$-th Gaussian, respectively. 
The SH coefficients comprise a DC component $\mathbf{h}_i^{\mathrm{dc}}$ and higher-order components $\mathbf{h}_i^{\mathrm{rest}}$, which describe the appearance of Gaussian primitives under different viewing directions. 
Given a camera pose $\pi$, the differentiable renderer produces an image $\mathbf{I}=\mathcal{R}(\mathcal{G},\pi)$. 
For dynamic Gaussian assets, temporal deformation transforms the Gaussian primitives over time, and the rendered image is expressed as $\mathbf{I}_t=\mathcal{R}(\mathcal{G},\pi,t)$~\cite{wu20244d}.

\subsection{Semantic-Interference-Aware Decoder Pretraining}
\label{sec:semantic_decoder}

\textbf{Decoder learning under semantic interference.}
To alleviate the semantic discrepancy between decoder pretraining and watermark extraction, we introduce a semantic-interference-aware pretraining strategy that incorporates scene-relevant visual semantics into decoder learning. 
Given a binary message $\mathbf{m}\in\{0,1\}^{L}$, where $L$ denotes the message length, we first convert the message into a CLIP-compatible token sequence $\mathbf{t}_{\mathbf{m}}$ using a predefined binary-to-token mapping. The frozen CLIP text encoder $E_T$ then extracts the corresponding message-related representation:
\begin{equation}
\mathbf{F}_{T}=E_T(\mathbf{t}_{\mathbf{m}}).
\end{equation}
To introduce visual semantic interference, an auxiliary image $\mathbf{I}_{s}$ is selected from a semantic image pool and encoded using the frozen CLIP visual encoder $E_V$:
\begin{equation}
\mathbf{F}_{I}=E_V(\mathbf{I}_{s}).
\end{equation}
The semantic image pool is constructed to expose the decoder to the visual content that may interfere with watermark extraction in Stage II. 
In our primary setting, the pool contains unwatermarked images from the scenes corresponding to the target Gaussian assets, providing semantic coverage of these assets in CLIP's visual feature space.
The images are included without explicit semantic similarity filtering or message-specific selection, and their content is independent of the binary watermark labels. We further consider an expanded pool that additionally incorporates images from scenes not involved in the corresponding watermark embedding experiments. This mixed-source setting allows us to investigate whether introducing additional non-target scene semantics improves or degrades the fidelity--recoverability trade-off. In both settings, all images are processed by the same frozen CLIP visual encoder used during Stage-II watermark extraction.

We combine the message-related textual feature and the auxiliary visual feature in the shared CLIP embedding space:
\begin{equation}
\mathbf{F}_{mix}=(1-\rho)\mathbf{F}_{T}+\rho\mathbf{F}_{I},
\label{eq:semantic_fusion}
\end{equation}
where $\rho\in[0,1)$ is the nominal mixing coefficient assigned to the visual feature. When $\rho=0$, the decoder receives only the textual watermark representation, corresponding to text-only decoder pretraining. 
As $\rho$ increases, a larger nominal weight is assigned to the visual feature, introducing stronger scene-dependent semantic interference into the mixed representation.
The resulting feature is normalized before being passed to the lightweight message decoder $D_{\psi}$:
\begin{equation}
\mathbf{o}=D_{\psi}\left(\frac{\mathbf{F}_{mix}}{\|\mathbf{F}_{mix}\|_2}\right),
\label{eq:decoder_forward}
\end{equation}
where $\mathbf{o}\in\mathbb{R}^{L}$ denotes the predicted message logits. The decoder is optimized using the binary cross-entropy loss:
\begin{equation}
\mathcal{L}_{dec}=\operatorname{BCE}(\mathbf{o},\mathbf{m}),
\label{eq:decoder_loss}
\end{equation}
where $\operatorname{BCE}$ denotes binary cross-entropy computed directly from the predicted logits and averaged over the message bits and training samples.
In SemGuard, semantic decoupling refers to reducing the influence of message-irrelevant scene semantics on watermark predictions, rather than explicitly factorizing CLIP features into separate semantic and watermark subspaces. Since the auxiliary visual features do not determine the target message, learning to recover messages from the mixed representations encourages the decoder to rely on message-related information while becoming less sensitive to scene-dependent variations.

\textbf{Progressive semantic interference curriculum.}
Directly introducing strong visual interference may make message recovery difficult before the decoder has established a reliable mapping between textual representations and binary messages. To facilitate decoder learning, we adopt a progressive semantic interference curriculum that gradually increases the visual mixing coefficient throughout pretraining. Specifically, at epoch $e$, the interference coefficient is defined as:
\begin{equation}
\rho_e=\rho_{\min}+(\rho_{\max}-\rho_{\min})\frac{e}{E},
\label{eq:rho_schedule}
\end{equation}
where $E$ denotes the total number of training epochs and $e\in\{1,\ldots,E\}$. 
The coefficient $\rho_e$ replaces $\rho$ in Eq.~\eqref{eq:semantic_fusion}. 
This schedule gradually shifts decoder learning from text-dominated representations toward stronger visual interference, facilitating message recovery under increasingly challenging semantic variations.
After pretraining for a given message length and semantic training distribution, the decoder is frozen and reused across multiple Gaussian assets.


\begin{table*}[!t]
\centering
\caption{Quantitative comparison of watermark recoverability and rendering fidelity on Blender and LLFF. $\star$ denotes SemGuard decoder pretraining with static scene images from Blender and LLFF, while $\dagger$ additionally incorporates dynamic scene images from D-NeRF and HyperNeRF. Bold indicates the best result for each metric.}
\label{tab:sota_comparison}
\resizebox{1.\linewidth}{!}{
\begin{tabular}{lcccccccccccc}
\toprule
\multirow{2}{*}{Methods} 
& \multicolumn{4}{c}{\textbf{16 bits}} 
& \multicolumn{4}{c}{\textbf{32 bits}} 
& \multicolumn{4}{c}{\textbf{48 bits}} \\
\cmidrule(lr){2-5} \cmidrule(lr){6-9} \cmidrule(lr){10-13}
& Bit Acc. $\uparrow$ & PSNR $\uparrow$ & SSIM $\uparrow$ & LPIPS $\downarrow$
& Bit Acc. $\uparrow$ & PSNR $\uparrow$ & SSIM $\uparrow$ & LPIPS $\downarrow$
& Bit Acc. $\uparrow$ & PSNR $\uparrow$ & SSIM $\uparrow$ & LPIPS $\downarrow$ \\
\midrule
\rowcolor{gray!15} \multicolumn{13}{l}{\textit{NeRF-based Watermarking Methods}} \\

CopyRNeRF~\cite{luo2023copyrnerf}
& 91.16 & 26.29 & 0.9100 & 0.0380 
& 78.08 & 26.13 & 0.8960 & 0.0410 
& 60.06 & 27.56 & 0.8950 & 0.0660 \\
WateRF~\cite{jang2024waterf}
& 95.67 & 32.79 & 0.9480 & 0.0330  
& 88.58 & 31.19 & 0.9360 & 0.0400 
& 85.82 & 30.86 & 0.9300 & 0.0400 \\   

\midrule
\rowcolor{gray!15} \multicolumn{13}{l}{\textit{3DGS Built on Watermarked Images}} \\
CIN~\cite{ma2022towards} + 3DGS~\cite{kerbl20233d}  
& 56.73 & 31.95 & 0.9194 & 0.1027 
& 53.13 & 31.74 & 0.9279 & 0.0944 
& 55.78 & 30.25 & 0.9139 & 0.1120 \\
SSL~\cite{fernandez2022watermarking} + 3DGS~\cite{kerbl20233d}    
& 58.94 & 36.51 & 0.9737 & 0.0094 
& 61.85 & 35.24 & 0.9706 & 0.0179 
& 58.79 & 35.88 & 0.9710 & 0.0123 \\
HiDDeN~\cite{zhu2018hidden} + 3DGS~\cite{kerbl20233d}
& 63.07 & 31.59 & 0.9790 & 0.0171 
& 52.46 & 34.51 & 0.9682 & 0.0209 
& 53.08 & 33.42 & 0.9687 & 0.0299 \\
DwtDctSvd~\cite{navas2008dwt} + 3DGS~\cite{kerbl20233d}
& 55.44 & 34.78 & 0.9582 & 0.0399 
& 53.15 & 32.32 & 0.9477 & 0.0547 
& 51.83 & 31.09 & 0.9302 & 0.0704 \\
StegaStamp~\cite{tancik2020stegastamp} + 3DGS~\cite{kerbl20233d}
& 79.72 & 35.52 & 0.9697 & 0.0181 
& 82.36 & 34.04 & 0.9601 & 0.0265 
& 83.97 & 32.54 & 0.9523 & 0.0406 \\

\midrule
\rowcolor{gray!15} \multicolumn{13}{l}{\textit{Gaussian Representation-based Watermarking}} \\

3DGS~\cite{kerbl20233d} w/ WateRF~\cite{jang2024waterf}
& 92.89 & 31.01 & 0.9678 & 0.0475 
& 90.15 & 29.56 & 0.9611 & 0.0492 
& 87.30 & 29.13 & 0.9562 & 0.0534 \\

3D-GSW~\cite{jang20253d}
&98.45 &36.07 &0.9834 & 0.0226 
& 98.01 & 36.16 & 0.9811 & 0.0422
& 94.13 & 33.42 & 0.9821 & 0.0431 \\

GaussianMarker~\cite{huang2024gaussianmarker}
& 99.36 & 34.42 & 0.9822 & 0.0124 
& 98.85 & 33.98 & 0.9788 & 0.0163 
& 98.25 & 32.12 & 0.9723 & 0.0234 \\

GuardSplat~\cite{chen2025guardsplat}
& 99.64 & 41.55 & 0.9957 & 0.0017
& 99.04 & 39.40 & 0.9939 & 0.0022 
& 98.29 & 38.90 & 0.9923 & 0.0028 \\

\midrule
\rowcolor{gray!15} \multicolumn{13}{l}{\textit{SemGuard with Different Pretraining Semantic Image Pools}} \\

\textbf{SemGuard(Ours$\dagger$)}
& 99.91 & 46.97 & 0.9986 & 0.0007
& \textbf{99.52} & \textbf{43.72} & \textbf{0.9975} & \textbf{0.0011}
& 98.05 & \textbf{42.58} & \textbf{0.9969} & \textbf{0.0014} \\

\textbf{SemGuard(Ours$\star$)}
& \textbf{99.97} & \textbf{50.29} & \textbf{0.9993} & \textbf{0.0004}
& 99.47 & 43.59 & \textbf{0.9975} & \textbf{0.0011}
& \textbf{98.72} & 42.32 & \textbf{0.9969} & \textbf{0.0014} \\

\bottomrule
\end{tabular}
}
\end{table*}

\subsection{SH-offset Gaussian Watermark Embedding}
\label{sec:embedding}

Following the SH-offset parameterization of GuardSplat~\cite{chen2025guardsplat}, SemGuard embeds the target message by optimizing additive perturbations to Gaussian SH coefficients while keeping the original geometry and deformation parameters fixed.
Let $\mathcal{G}$ and $\widetilde{\mathcal{G}}$ denote the original and watermarked Gaussian assets, respectively. For each Gaussian primitive $i$, the watermarked SH coefficients are defined as:
\begin{equation}
\widetilde{\mathbf{h}}_i^k=\mathbf{h}_i^k+\Delta\mathbf{h}_i^k,\quad k\in\{\mathrm{dc},\mathrm{rest}\},
\label{eq:sh_offset}
\end{equation}
where $\Delta\mathbf{h}_i^k$ denotes the learnable appearance offset for the corresponding SH component. The offsets are collected into $\Delta\mathbf{H}^{\mathrm{dc}}$ and $\Delta\mathbf{H}^{\mathrm{rest}}$. During watermark embedding, only these SH offsets are optimized, while Gaussian positions, scales, rotations, opacity, and the deformation parameters of dynamic Gaussian assets remain fixed.

\textbf{Watermark extraction from static Gaussian assets.}
Given a camera pose $\pi$, we render the watermarked asset and its original counterpart as $\widetilde{\mathbf{I}}=\mathcal{R}(\widetilde{\mathcal{G}},\pi)$ and $\mathbf{I}=\mathcal{R}(\mathcal{G},\pi)$, respectively. The watermarked rendering is processed by the frozen CLIP visual encoder, and its visual feature is normalized before being passed to the frozen message decoder:
\begin{equation}
\mathbf{o}=D_{\psi}\left(\frac{E_V(\widetilde{\mathbf{I}})}{\|E_V(\widetilde{\mathbf{I}})\|_2}\right).
\label{eq:visual_decode}
\end{equation}
The message loss guides SH-offset optimization through differentiable rendering, allowing gradients to propagate through the frozen visual encoder and message decoder to the learnable appearance offsets.

\textbf{Watermark extraction from dynamic Gaussian assets.}
For dynamic Gaussian assets, watermark extraction must account for appearance variations induced by temporal deformation and changes in viewing conditions. We therefore adopt a simple temporal feature aggregation strategy before message decoding. Consider a sampled viewpoint sequence $v$ containing $J$ frames with camera poses $\{\pi_{v,j}\}_{j=1}^{J}$ and timestamps $\{t_j\}_{j=1}^{J}$. The corresponding watermarked frames are rendered as $\widetilde{\mathbf{I}}_{v,j}=\mathcal{R}(\widetilde{\mathcal{G}},\pi_{v,j},t_j)$. Their CLIP visual features are averaged before normalization and message decoding:
\begin{equation}
\begin{aligned}
\overline{\mathbf{F}}_{V,v} &= \frac{1}{J}\sum_{j=1}^{J}E_V\left(\widetilde{\mathbf{I}}_{v,j}\right),\\
\mathbf{o}_v &= D_{\psi}\left(\frac{\overline{\mathbf{F}}_{V,v}}{\|\overline{\mathbf{F}}_{V,v}\|_2}\right).
\end{aligned}
\label{eq:temporal_decode}
\end{equation}
Here, $\overline{\mathbf{F}}_{V,v}$ denotes the temporally aggregated visual feature of viewpoint sequence $v$. 
Unlike frame-wise decoding, this strategy produces one message prediction per viewpoint sequence, with the same target message supervised across sampled sequences. The aggregation operates directly in the frozen CLIP feature space.


\begin{table*}[!t]
\centering
\caption{Quantitative comparison of watermark recoverability and rendering fidelity on D-NeRF and HyperNeRF. $\triangle$ denotes decoder pretraining using dynamic scene images from D-NeRF and HyperNeRF, while $\dagger$ additionally includes static scene images from Blender and LLFF.}
\label{tab4D:4D}
\resizebox{1.\linewidth}{!}{
\begin{tabular}{lcccccccccccc}
\toprule
\multirow{2}{*}{Methods} 
& \multicolumn{4}{c}{\textbf{16 bits}} 
& \multicolumn{4}{c}{\textbf{32 bits}} 
& \multicolumn{4}{c}{\textbf{48 bits}} \\
\cmidrule(lr){2-5} \cmidrule(lr){6-9} \cmidrule(lr){10-13}
& Bit Acc. $\uparrow$ & PSNR $\uparrow$ & SSIM $\uparrow$ & LPIPS $\downarrow$
& Bit Acc. $\uparrow$ & PSNR $\uparrow$ & SSIM $\uparrow$ & LPIPS $\downarrow$
& Bit Acc. $\uparrow$ & PSNR $\uparrow$ & SSIM $\uparrow$ & LPIPS $\downarrow$ \\
\midrule
\rowcolor{gray!15} \multicolumn{13}{l}{\textit{Gaussian-based watermarking methods} } \\
GuardSplat~\cite{chen2025guardsplat}
& 87.61 & 43.82 & 0.9972 & 0.0014
& 76.06 & 40.35 & 0.9947 & 0.0029 
& 72.61 & 38.98 & 0.9935 & 0.0039 \\

Mark4D~\cite{lee2026mark4d}
& 88.27 & 43.96 & 0.9974 & 0.0013
& 75.20 & 41.39 & 0.9946 & 0.0030 
& 73.57 & 39.24 & 0.9934 & 0.0038 \\

\rowcolor{gray!15} \multicolumn{13}{l}{\textit{SemGuard with Different Pretraining Semantic Image Pools}} \\

\textbf{SemGuard(Ours$\dagger$)}
& \textbf{95.35} & \textbf{47.39} & \textbf{0.9987} & 0.0010
& \textbf{91.91} & \textbf{45.41} & \textbf{0.9981} & \textbf{0.0011}
& 86.85 & 43.31 & 0.9971 & \textbf{0.0015} \\

\textbf{SemGuard(Ours$\triangle$)}
& 92.75 & 47.23 & 0.9985 & \textbf{0.0008}
& 90.67 & 44.05 & 0.9975 & 0.0014
& \textbf{87.89} & \textbf{43.35} & \textbf{0.9972} & \textbf{0.0015} \\

\bottomrule
\end{tabular}
}
\end{table*}

\subsection{Optimization and Verification}
\label{sec:objective}

During watermark embedding, the pretrained CLIP visual encoder $E_V$ and message decoder $D_{\psi}$ remain fixed, while only the Gaussian SH offsets are optimized. 
The optimization objective balances watermark recoverability, preservation of the original rendering appearance, and the magnitude of the SH perturbations.

\textbf{Message recovery loss.}
For a mini-batch of $B$ decoded views or viewpoint sequences, we minimize the binary cross-entropy between the predicted message logits and the target binary message:
\begin{equation}
\mathcal{L}_{\mathrm{msg}}=\frac{1}{B}\sum_{b=1}^{B}\operatorname{BCE}(\mathbf{o}_b,\mathbf{m}),
\label{eq:msg_loss}
\end{equation}
where each $\mathbf{o}_b$ corresponds to an individual rendered view for static assets or a temporally aggregated viewpoint sequence for dynamic assets. The same target message $\mathbf{m}$ is supervised across all sampled observations.

\textbf{Rendering fidelity loss.}
To preserve rendering fidelity, we constrain the difference between the watermarked rendering $\widetilde{\mathbf{I}}$ and the corresponding rendering $\mathbf{I}$ of the original unwatermarked Gaussian asset at the same viewpoint and timestamp. 
We combine pixel-wise reconstruction, structural similarity, and perceptual consistency:
\begin{equation}
\begin{aligned}
\mathcal{L}_{\mathrm{recon}} &= (1-\lambda_{\mathrm{ssim}})\operatorname{MAE}(\widetilde{\mathbf{I}},\mathbf{I})\\
&\quad+\lambda_{\mathrm{ssim}}\left[1-\operatorname{SSIM}(\widetilde{\mathbf{I}},\mathbf{I})\right]\\
&\quad+\lambda_{\mathrm{lpips}}\operatorname{LPIPS}(\widetilde{\mathbf{I}},\mathbf{I}),
\end{aligned}
\label{eq:recon_loss}
\end{equation}
where $\operatorname{MAE}$ denotes the mean absolute pixel error. For dynamic assets, the rendering fidelity losses are evaluated on individual frames, whereas message recovery is supervised using temporally aggregated features. 

\textbf{SH-offset regularization.}
To discourage excessive appearance perturbations, we penalize the magnitudes of the optimized SH offsets:
\begin{equation}
\mathcal{L}_{\mathrm{off}}=\operatorname{MSE}(\Delta\mathbf{H}^{\mathrm{dc}},\mathbf{0})+\operatorname{MSE}(\Delta\mathbf{H}^{\mathrm{rest}},\mathbf{0}),
\label{eq:offset_loss}
\end{equation}
where $\operatorname{MSE}$ denotes the element-wise mean squared error. Combining the three loss terms, the final embedding objective is:
\begin{equation}
\mathcal{L}=\lambda_{\mathrm{msg}}\mathcal{L}_{\mathrm{msg}}+\lambda_{\mathrm{recon}}\mathcal{L}_{\mathrm{recon}}+\lambda_{\mathrm{off}}\mathcal{L}_{\mathrm{off}}.
\label{eq:full_objective}
\end{equation}
The objective is not to maximize Bit Accuracy at any visual cost, but to obtain reliable message extraction while keeping watermark-induced rendering distortion as small as possible. The message recovery term enforces watermark recoverability, whereas the reconstruction and offset regularization terms discourage unnecessary modifications to the original Gaussian appearance.

\textbf{Watermark extraction evaluation.}
After watermark embedding, the target message $\mathbf{m}$ is compared with the predicted messages extracted from rendered views. Let $\mathbf{o}_v$ denote the message logits from the $v$-th evaluated view or viewpoint sequence, obtained using Eq.~\eqref{eq:visual_decode} or Eq.~\eqref{eq:temporal_decode}, respectively. The predicted value of the $q$-th message bit is $\hat{m}_{v,q}=\mathbb{I}(o_{v,q}>0)$, where $\mathbb{I}$ denotes the indicator function. 
Given $K$ evaluated views or viewpoint sequences, the average Bit Accuracy is:
\begin{equation}
\mathcal{A}=\frac{100}{KL}\sum_{v=1}^{K}\sum_{q=1}^{L}\left(1-|\hat{m}_{v,q}-m_q|\right).
\label{eq:verify}
\end{equation}
Here, $\mathcal{A}$ represents the percentage of correctly recovered message bits, averaged across the evaluated views or viewpoint sequences. 
This metric measures watermark extraction accuracy for both static and dynamic Gaussian assets.

\begin{table*}[!t]
\centering
\caption{Ablation study of semantic interference strength and progressive training strategies on Blender and LLFF using scene images from the static assets. Fixed mixing coefficients $\rho\in\{0,0.1,0.3,0.5,0.7,0.9\}$ are compared with progressive interference schedules to evaluate their effects on rendering fidelity and watermark recoverability. }
\label{tab:semantic_ablation}
\resizebox{1.\linewidth}{!}{
\begin{tabular}{lcccccccccccc}
\toprule
\multirow{2}{*}{Strategy} 
& \multicolumn{4}{c}{\textbf{16 bits}} 
& \multicolumn{4}{c}{\textbf{32 bits}} 
& \multicolumn{4}{c}{\textbf{48 bits}} \\
\cmidrule(lr){2-5} \cmidrule(lr){6-9} \cmidrule(lr){10-13}
& Bit Acc. $\uparrow$ & PSNR $\uparrow$ & SSIM $\uparrow$ & LPIPS $\downarrow$
& Bit Acc. $\uparrow$ & PSNR $\uparrow$ & SSIM $\uparrow$ & LPIPS $\downarrow$
& Bit Acc. $\uparrow$ & PSNR $\uparrow$ & SSIM $\uparrow$ & LPIPS $\downarrow$ \\
\midrule

\rowcolor{gray!15} \multicolumn{13}{l}{\textit{Fixed interference weights}} \\
0
& 99.64 & 41.55 & 0.9957 & 0.0017
& 99.04 & 39.40 & 0.9939 & 0.0022 
& 98.29 & 38.90 & 0.9923 & 0.0028 \\

0.1
& 99.95 & 47.10 & 0.9987 & \textbf{0.0004}
& 98.17 & 39.91 & 0.9942 & 0.0028 
& 95.81 & 38.25 & 0.9923 & 0.0036 \\

0.3
& 99.93 & 46.27 & 0.9984 & 0.0008
& 98.02 & 40.58 & 0.9951 & 0.0023  
& 98.12 & 40.25 & 0.9950 & 0.0024 \\

0.5
& 99.96 & 48.23 & 0.9990 & 0.0005
& 98.80 & 41.14 & 0.9958 & 0.0020 
& 98.59 & 40.66 & 0.9955 & 0.0021 \\

0.7
& 99.96 &50.10 & \textbf{0.9993} & \textbf{0.0004}
& 99.21 & 42.59 & 0.9967 & 0.0015 
& 98.70 & 41.88 & 0.9963 & 0.0017 \\

0.9
& 99.89 & 45.45  & 0.9981 & 0.0010
& 93.11 & 43.16 & 0.9972 & 0.0012 
& 86.42 & 41.84 & 0.9967 & 0.0018 \\

\rowcolor{gray!15} \multicolumn{13}{l}{\textit{Progressive interference schedules}} \\
Step
& 99.96 & 48.10 & 0.9989 & 0.0005
& 99.18 & 42.59 & 0.9969 & 0.0014 
& 98.37 & 42.23 & 0.9968 & \textbf{0.0014} \\

Continuous 
& \textbf{99.97} & \textbf{50.29} & \textbf{0.9993} & \textbf{0.0004}
& \textbf{99.47} & \textbf{43.59} & \textbf{0.9975} & \textbf{0.0011} 
& \textbf{98.72} & \textbf{42.32} & \textbf{0.9969} & \textbf{0.0014} \\

\bottomrule
\end{tabular}
}
\end{table*}

\begin{table*}[!t]
\centering
\caption{Bit Accuracy (\%) under image-space distortions for 16-bit 3DGS watermarking. $\star$ denotes decoder pretraining with images from Blender and LLFF, while $\dagger$ additionally includes images from D-NeRF and HyperNeRF.}
\label{tab:distortion_comparison}
\resizebox{1.\linewidth}{!}{
\begin{tabular}{lcccccccccc}
\toprule
\textbf{Methods}
& \multicolumn{10}{c}{\textbf{Bit Acc. (\%) $\uparrow$}} \\
\cmidrule(lr){2-11}
& \textbf{None}
& \textbf{\begin{tabular}[c]{@{}c@{}}Noise\\$(\sigma{=}0.1)$\end{tabular}}
& \textbf{\begin{tabular}[c]{@{}c@{}}Rotation\\$(\pm\pi/6)$\end{tabular}}
& \textbf{\begin{tabular}[c]{@{}c@{}}Scaling\\$(\leq25\%)$\end{tabular}}
& \textbf{\begin{tabular}[c]{@{}c@{}}Blur\\$(\sigma{=}0.1)$\end{tabular}}
& \textbf{\begin{tabular}[c]{@{}c@{}}Crop\\$(40\%)$\end{tabular}}
& \textbf{\begin{tabular}[c]{@{}c@{}}Brightness\\$(0.5{\sim}1.5)$\end{tabular}}
& \textbf{\begin{tabular}[c]{@{}c@{}}JPEG\\$(10\%\ \text{quality})$\end{tabular}}
& \textbf{\begin{tabular}[c]{@{}c@{}}VAE Attack~\cite{zhao2024invisible}\end{tabular}}
& \textbf{Combined} \\
\midrule  
CopyRNeRF~\cite{luo2023copyrnerf}
& 91.16 & 90.04 & 88.13 & 89.33 & 90.06 & -- & -- & -- & -- & -- \\

WateRF~\cite{jang2024waterf}
& 95.67 & 95.36 & 93.13 & 93.29 & 95.25 & 95.40 & 90.91 & 86.99 & 51.73 & 84.12 \\

3DGS~\cite{kerbl20233d} w/ WateRF~\cite{jang2024waterf}
& 92.89 & 87.35 & 88.28 & 90.33 & 91.92 & 89.07 & 88.71 & 88.49 & 55.48 & 86.37 \\

GaussianMarker~\cite{huang2024gaussianmarker}
& 99.36 & 99.13 & 70.84 & 97.89 & 94.40 & 98.52 & 95.78 & 86.22 & 52.00 & 83.49 \\

3D-GSW~\cite{jang20253d} & 98.36 & 84.95 & 88.14 & 94.98 & 96.43 & 93.18 &94.68 & 93.71 &67.23 & 91.13 \\

GuardSplat~\cite{chen2025guardsplat}
&99.64 &99.60 &94.56 &98.75 &99.27 &98.71 &97.46 &94.70 &82.35 & 93.38 \\

\textbf{SemGuard(Ours$\dagger$)} 
&99.91 &99.79 &96.86 &99.56 &99.90 &99.24 &98.39 &96.48 &95.04 & 96.65 \\

\textbf{SemGuard(Ours$\star$ )}     
&99.97 &99.81 &97.01 &99.98 &99.94 &99.15 &99.40 &96.53 &95.26 & 97.24 \\

\bottomrule
\end{tabular}
}
\end{table*}

\section{Experiments}

\subsection{Datasets and Evaluation Protocol}
We evaluate SemGuard on both static and dynamic Gaussian assets. For 3DGS watermarking, we use Blender~\cite{mildenhall2021nerf} and LLFF~\cite{mildenhall2019local}. For 4DGS watermarking, we use D-NeRF~\cite{pumarola2021d} and HyperNeRF~\cite{park2021hypernerf}. 
During Stage-I pretraining, we construct semantic image pools from unwatermarked scene images to provide visual semantic coverage of the Gaussian assets subsequently processed in Stage II. We investigate three configurations: a static pool containing Blender and LLFF images ($\star$), a dynamic pool containing D-NeRF and HyperNeRF images ($\triangle$), and a mixed pool containing images from all four datasets ($\dagger$). For static watermarking, the static pool provides the target-scene semantic coverage, whereas the mixed pool additionally introduces images from dynamic scenes that are not watermarked in the corresponding experiments. For dynamic watermarking, the roles are reversed: the dynamic pool covers the target scenes, and the mixed pool additionally incorporates static-scene images. These configurations are designed to examine the influence of additional non-target scene semantics rather than to establish scene-disjoint generalization. For each message length and image pool, a single decoder is pretrained and frozen for reuse across all corresponding Gaussian assets, while SH offsets are optimized independently for each asset. In the evaluated protocol, the target watermark message is sampled from the message set used during decoder pretraining, while watermark embedding is performed independently for each Gaussian asset.

\subsection{Evaluation Metrics}
\label{subsec:metrics}

We evaluate Gaussian watermarking performance by jointly considering rendering fidelity and watermark extraction reliability, with particular emphasis on preserving the original rendered appearance while ensuring reliable message recovery. To quantify watermark-induced visual distortions rather than reconstruction errors relative to dataset ground-truth images, we report PSNR, SSIM~\cite{wang2004image}, and LPIPS~\cite{zhang2018unreasonable} between watermarked and original unwatermarked Gaussian renderings at matching camera viewpoints and timestamps. Higher PSNR and SSIM, together with lower LPIPS, indicate better preservation of the original rendered appearance. Watermark extraction reliability is measured by Bit Accuracy (Bit Acc., \%), which represents the proportion of correctly recovered message bits. For 3DGS assets, each rendered viewpoint is decoded independently, whereas for 4DGS assets, CLIP visual features are temporally averaged within each viewpoint sequence before message decoding, and Bit Acc. is subsequently averaged across the evaluated sequences. We further assess watermark extraction robustness against image-space distortions on static assets, including noise, rotation, scaling, blur, cropping, brightness changes, JPEG compression, VAE-based attacks, and combined distortions.

\subsection{Implementation Details}
We use CLIP ViT-B/32~\cite{radford2021learning} as the frozen visual and textual encoder.
For Stage-I decoder learning, only the lightweight message decoder is optimized.
For Stage-II watermark embedding, the decoder and CLIP image encoder are frozen, and only SH offsets are optimized while keeping Gaussian geometry, opacity, and deformation parameters fixed.
We train decoders for different message lengths separately and optimize SH offsets with Adam.
The detailed training configurations and hyperparameters are provided in the supplementary material.

\subsection{Comparison with Other Methods}
\label{subsec:comparison}

\textbf{Results on 3DGS assets.}
Table~\ref{tab:sota_comparison} compares watermark extraction and rendering fidelity on Blender and LLFF with 16-, 32-, and 48-bit messages. Among methods employing Gaussian representation-based watermarking, GuardSplat is particularly relevant because it also uses a pretrained CLIP-guided decoder and SH-based watermark embedding. 
The most significant advantage of SemGuard lies in preserving rendering fidelity while maintaining highly reliable watermark extraction.  
For 16-bit watermarking on Blender and LLFF, GuardSplat already achieves a near-saturated Bit Accuracy of 99.64\%, leaving limited room for improvement in extraction accuracy alone.
Nevertheless, SemGuard($\star$) increases PSNR from 41.55 dB to 50.29 dB, improves SSIM from 0.9957 to 0.9993, and reduces LPIPS from 0.0017 to 0.0004, while achieving a Bit Accuracy of 99.97\%.
These results indicate that the semantic-interference-aware decoder enables the SH-offset embedding procedure to preserve the original rendered appearance substantially better without sacrificing watermark recoverability.
At 32 bits, SemGuard($\dagger$) improves PSNR from 39.40 dB to 43.72 dB relative to GuardSplat, while Bit Acc. increases from 99.04\% to 99.52\%. At 48 bits, SemGuard($\star$) achieves 42.32 dB PSNR and 98.72\% Bit Acc., compared with 38.90 dB and 98.29\% for GuardSplat. The semantic image pool also affects the outcome. Although the mixed-source variant yields the highest 48-bit PSNR (42.58 dB), its Bit Acc. (98.05\%) is below that of GuardSplat (98.29\%) and SemGuard($\star$) (98.72\%). 
These results suggest that the effectiveness of a semantic image pool may depend on how well its visual content represents the semantic interference encountered during subsequent watermark extraction. Incorporating images from non-target scenes does not necessarily improve the balance between rendering fidelity and watermark recoverability.

\textbf{Results on 4DGS assets.}
Table~\ref{tab4D:4D} evaluates dynamic watermarking on D-NeRF and HyperNeRF using viewpoint-sequence predictions; the reported aggregates assign equal weight to the two benchmark datasets. This setting is particularly relevant because motion and temporal appearance changes modify the visual evidence available to the decoder. 
At 32 bits, SemGuard($\dagger$) substantially improves rendering fidelity, achieving 45.41 dB PSNR, 0.9981 SSIM, and 0.0011 LPIPS, compared with 40.35 dB, 0.9947, and 0.0029 for GuardSplat. Meanwhile, Bit Acc. increases from 76.06\% to 91.91\%, indicating that the improved visual fidelity is accompanied by more reliable watermark recovery. Compared with Mark4D, SemGuard($\dagger$) also achieves better results in both rendering fidelity and extraction accuracy under the evaluated protocol.
At 48 bits, SemGuard($\triangle$) achieves 87.89\% Bit Acc. and 43.35 dB PSNR, compared with 86.85\% and 43.31 dB for SemGuard($\dagger$). These results indicate that incorporating additional static-scene images during decoder pretraining does not uniformly improve message recovery on dynamic assets, while the two variants maintain similar rendering fidelity. 
Since the compared methods employ different decoding and temporal-processing designs, the table establishes the performance of their complete watermarking pipelines rather than isolating the effect of decoder pretraining.

\begin{figure*}[t!]
	\centering
 \includegraphics[width=7in]{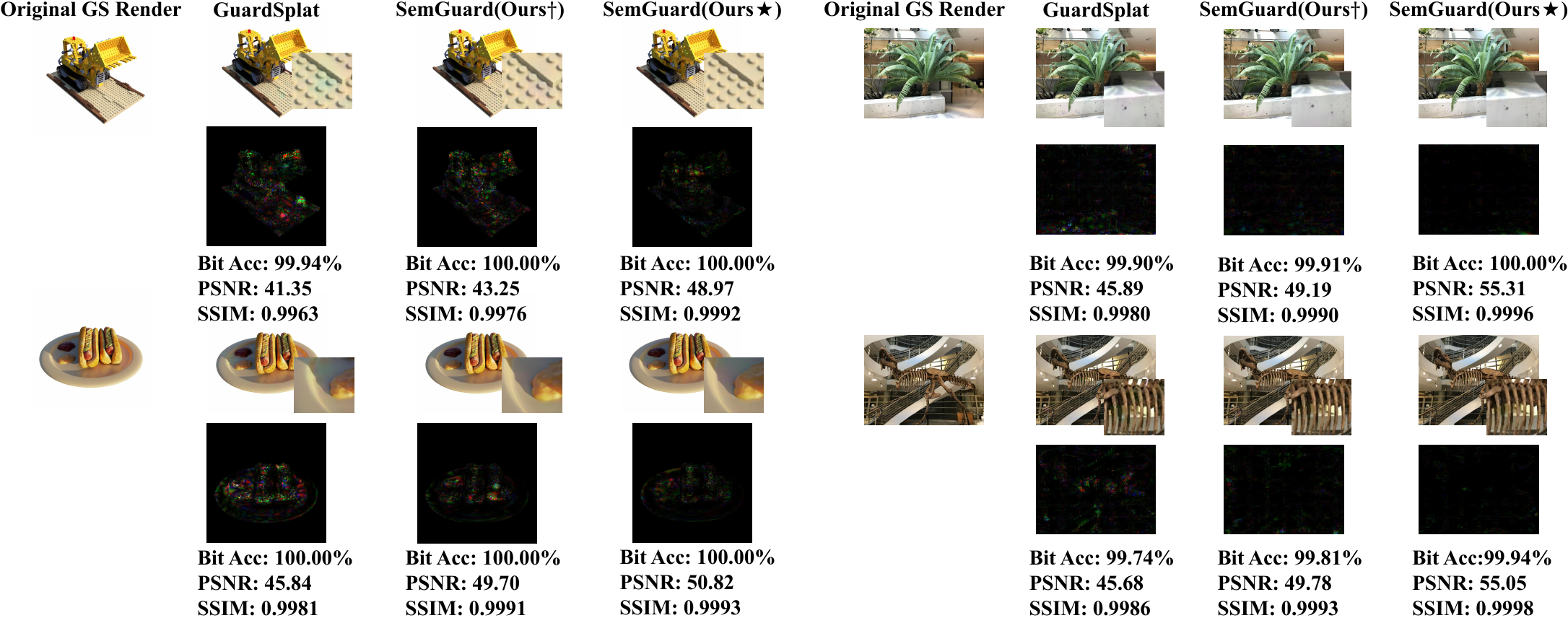}
\caption{Qualitative comparison of 16-bit watermarking on Blender and LLFF. Error maps visualize the absolute differences between watermarked and original Gaussian renderings, amplified by a factor of 10 for better visibility, with darker regions indicating smaller deviations.}
\label{fig:view}
\end{figure*}

\subsection{Ablation Study}

\textbf{Effect of semantic interference and progressive curriculum.}
Table~\ref{tab:semantic_ablation} investigates the effects of scene-relevant visual interference and its weighting strategy on 3DGS watermarking. 
The $\rho=0$ setting represents text-only decoder pretraining and is included as a reference baseline for evaluating the effect of incorporating visual semantic interference. The comparison with progressive training illustrates the resulting differences in rendering fidelity and watermark recoverability. 
Compared with this baseline, the continuous curriculum improves PSNR from 41.55 dB to 50.29 dB at 16 bits, from 39.40 dB to 43.59 dB at 32 bits, and from 38.90 dB to 42.32 dB at 48 bits, while maintaining or improving watermark extraction accuracy.
These results indicate that incorporating scene-relevant visual semantics during decoder pretraining can improve the fidelity--recoverability trade-off under the evaluated SH-offset embedding protocol, while the final performance depends on the interference strength and its training schedule.
The effect of a fixed interference coefficient is non-monotonic. At 32 bits, increasing $\rho$ from 0.7 to 0.9 improves PSNR from 42.59 dB to 43.16 dB but reduces Bit Acc. from 99.21\% to 93.11\%, highlighting the importance of balancing rendering fidelity and watermark recoverability.
 At 48 bits, the continuous curriculum achieves 98.72\% Bit Acc. and 42.32 dB PSNR, compared with 86.42\% and 41.84 dB under fixed $\rho=0.9$. Although fixed $\rho=0.7$ remains competitive at 48 bits, achieving 98.70\% Bit Acc. and 41.88 dB PSNR, the continuous curriculum provides a more favorable fidelity--recoverability trade-off across the evaluated message lengths. These results demonstrate the benefit of incorporating scene-relevant visual semantics during decoder pretraining, with progressive weighting further improving the balance between rendering fidelity and reliable watermark recovery.

\subsection{Robustness against Distortions}
\label{subsec:distortion}
Table~\ref{tab:distortion_comparison} compares 16-bit watermark extraction accuracy under image-space distortions on Blender and LLFF. These experiments assess robustness to rendered-image modifications rather than attacks on Gaussian model parameters. Under the VAE-based attack, SemGuard($\star$) retains 95.26\% Bit Acc., compared with 82.35\% for GuardSplat; the decrease relative to clean extraction is 4.71 and 17.29 percentage points, respectively. SemGuard($\star$) also achieves 97.01\% under rotation and 96.53\% under JPEG compression, compared with 94.56\% and 94.70\% for GuardSplat. 
These results demonstrate that SemGuard maintains more reliable watermark extraction than GuardSplat under the evaluated image-space distortions. Furthermore, SemGuard($\star$) generally outperforms SemGuard($\dagger$) on the static assets, including under combined attacks (97.24\% versus 96.65\%), indicating that introducing additional semantic sources does not necessarily improve robustness for the target Gaussian scenes.


\subsection{Qualitative Comparison}

Figure~\ref{fig:view} presents qualitative comparisons on representative Blender and LLFF scenes. 
While GuardSplat and both SemGuard variants achieve near-saturated 16-bit watermark extraction accuracy, SemGuard($\star$) better preserves the original rendered appearance, as reflected by the reduced differences in the error maps and magnified insets. 
These visual observations are consistent with its higher PSNR and SSIM, demonstrating improved rendering fidelity without compromising watermark recovery.
Compared with the mixed-source variant, SemGuard($\star$) also produces smaller rendering deviations in the illustrated scenes, suggesting that incorporating additional dynamic-scene semantics during decoder pretraining does not necessarily improve rendering fidelity for the target static Gaussian assets. 

\section{Conclusion}
\label{sec:conclusion}

We presented SemGuard, a simple yet effective semantics-decoupled watermarking framework that incorporates scene-relevant visual semantics from unwatermarked images into universal message decoder pretraining. 
By progressively increasing the contribution of scene-dependent visual features, SemGuard encourages the decoder to reduce its sensitivity to semantic interference while preserving reliable watermark recovery. 
The pretrained decoder remains frozen during SH-offset watermark embedding and can be reused across multiple Gaussian assets whose visual semantics are covered by the pretraining image pool. 
Experiments on static and dynamic Gaussian Splatting assets demonstrate substantial improvements in rendering fidelity, as measured by PSNR, SSIM, and LPIPS, while maintaining reliable watermark extraction. Additional evaluations on static Gaussian assets further demonstrate enhanced robustness against image-space distortions.
The results further suggest that watermarking performance depends on the semantic coverage and composition of the pretraining image pool, as well as the interference training schedule. 
Future work will explore scaling decoder pretraining to larger image datasets spanning diverse semantic distributions, with the aim of improving generalization across a broader range of Gaussian assets.

\section*{Acknowledgments}
This work was supported in part by the National Natural Science Foundation of China (U25A20441), the National Science and Technology Major Project of China (2025ZD1700704), the National Key Research and Development Program of China (2024YFB3311703), and Zhongguancun Laboratory.

\bibliographystyle{IEEEtran}
\bibliography{reference.bib}


\vfill

\end{document}